%% file: root.tex
\documentclass[letterpaper, 10 pt, conference]{ieeeconf}  

\usepackage{graphicx}
\IEEEoverridecommandlockouts                              

\title{\LARGE \bf
Contrastive Learning for Multimodal Human Activity Recognition with Limited Labeled Data
}

\author{Long Jing$^{1}$, Zhixiong Yang$^{2*}$, Yajun Zhang$^{2}$, and Xinlong Feng$^{2*}$
\thanks{*Corresponding authors.}
\thanks{This work was partially supported by the National Natural Science Foundation of China (Grant Nos. 12271465), and the Natural Science Foundation of Xinjiang Province (Grant Nos. 2022TSYCTD0019 and 2022D01D32).}%
\thanks{$^{1}$Long Jing is with the College of Computer Science, Northwest University, Xi'an, China 
        {\tt\small 2023117285@stumail.nwu.edu.cn}}%
\thanks{$^{2}$Zhixiong Yang, Yajun Zhang, and Xinlong Feng are with Xinjiang University, Urumqi, China 
        {\tt\small \{yzx, zyj, fxlmath\}@xju.edu.cn}}%
}
\usepackage{tabularx} 
\usepackage{booktabs} 
\usepackage{amssymb}  
\usepackage{amsmath}

\begin{document}

\maketitle
\thispagestyle{empty}
\pagestyle{empty}

\begin{abstract}

Human activity recognition serves as the foundation for various emerging applications. In recent years, researchers have used collaborative sensing of multi-source sensors to capture complex and dynamic human activities. However, multimodal human activity sensing typically encounters highly heterogeneous data across modalities and label scarcity, resulting in an application gap between existing solutions and real-world needs. In this paper, we propose CLMM, a general contrastive learning framework for human activity recognition that achieves effective multimodal recognition with limited labeled data. 
CLMM employs a novel two-stage training strategy. In the first stage, CLMM employs a CNN-DiffTransformer encoder to capture cross-modal shared information by extracting local and global features. Meanwhile, a hard-positive samples weighting algorithm enhances gradient propagation to reinforce shared learning. In the second stage, a dual-branch architecture combining quality-guided attention and bidirectional gated units captures modality-specific information, while a primary–auxiliary collaborative training strategy fuses both shared and modality-specific information. Experimental results on three public datasets demonstrate that CLMM significantly improves state-of-the-art baselines in both recognition accuracy and convergence performance.

\end{abstract}


\input{1.Introduction}
\input{2.Related_work}

\input{5.System_design}

\input{6.implement}
\input{7.evaluation}

\input{9.conclusion}

\addtolength{\textheight}{-12cm}   









\bibliographystyle{IEEEtran}
\bibliography{reference}

\end{document}

%% file: 1.introduction.tex
\section{Introduction}

With the rapid development of mobile sensing and IoT technologies, human activity recognition (HAR) has become fundamental to applications such as virtual reality~\cite{pascoal2019activity,rabbi2018virtual}, smart homes~\cite{adaimi2021ok,bi2017familylog}, and health monitoring~\cite{chen2024exploring,guo2024exploring}.
Recent studies increasingly use multiple sensors to capture complementary information about human motion, posture, and context, enabling more comprehensive recognition of complex activities~\cite{dai2024advancing,xue2019deepfusion}.
For example, in sports medicine, combining foot pressure sensors with IMUs can monitor gait freezing in Parkinson’s patients in real time~\cite{kourtis2019digital}.


However, fusing multiple sensor modalities in HAR still presents several challenges.
First, data from different sensors are highly heterogeneous in sampling frequency, scale, and structure, which makes cross-modal alignment difficult and weakens generalization~\cite{haresamudram2021contrastive}.
Second, high annotation costs and limited labeled samples are common issues in  multimodal HAR~\cite{hu2020fine,shuai2021millieye}, as annotating multimodal data not only requires synchronising information across different modalities but also demands that annotators possess specialised knowledge in the multimodal field.
Third, sensor data are inherently privacy-sensitive, which prevents large-scale annotated data from being uploaded to the cloud for centralized training.
Finally, HAR models trained in the cloud may require customization during local deployment, since individual activity patterns can change dynamically over time. This necessitates continuous multimodal training on local devices, which places high demands on hardware resources and significantly increases deployment costs~\cite{ouyang2021clusterfl}.

 \begin{figure}[!t] 
	\centering
	\includegraphics[width=3.4in]{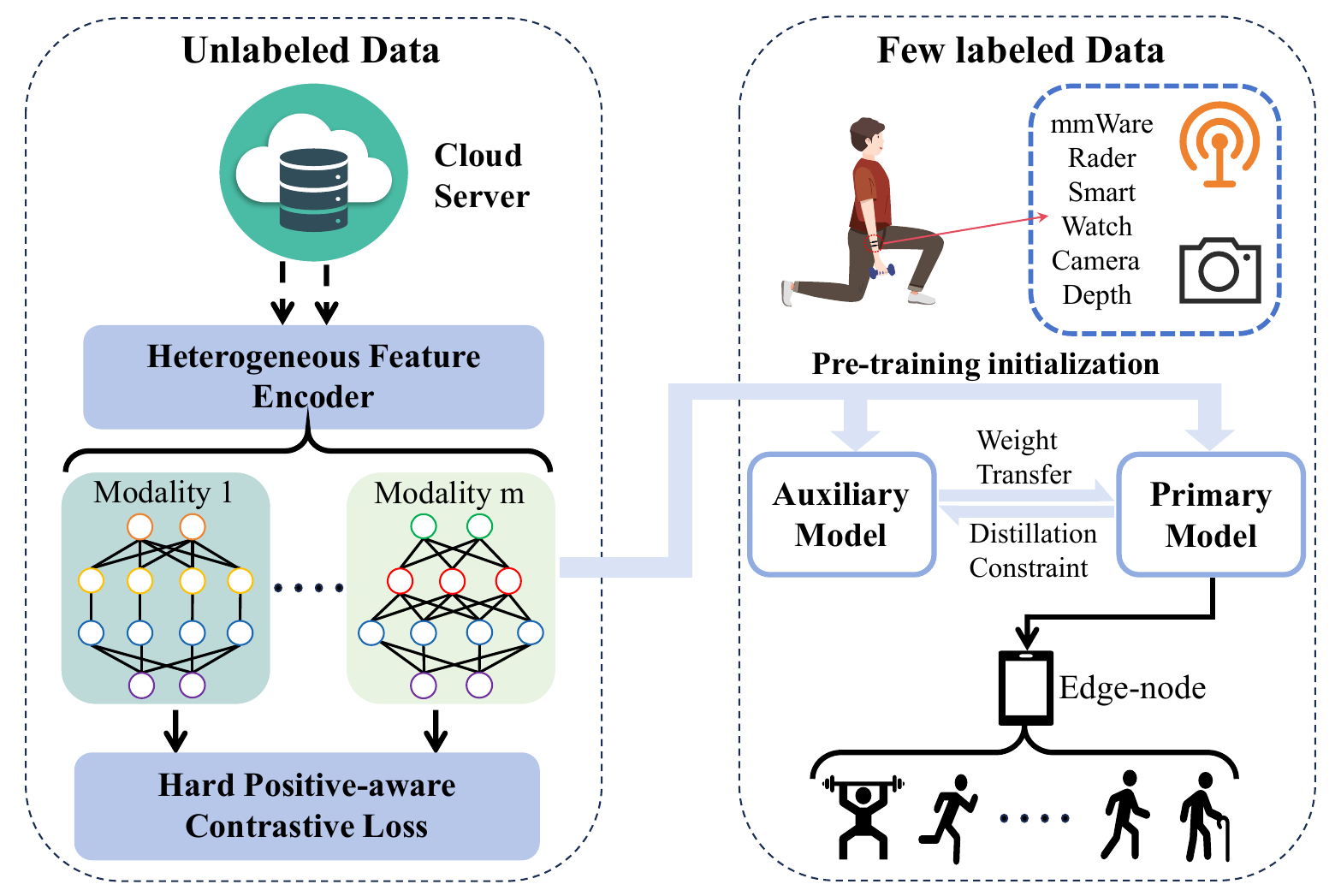}
	\caption{A typical application of CLMM in multimodal human activity recognition.}
	\label{fig1}
\end{figure}

Unfortunately, these challenges remain difficult to address simultaneously in practical HAR systems.
Current systems are either based on supervised learning frameworks for multimodal HAR or on self-supervised strategies tailored to modalities with similar characteristics. For example, RF-Camera integrates vision techniques with RFID scheduling to recognize gesture interactions in multi-subject and multi-target scenarios. However, it requires large amounts of annotated data, making it unsuitable for real-world HAR applications with limited labeled data~\cite{liu2021rfid}. CPCHAR employs a contrastive predictive coding (CPC) framework~\cite{haresamudram2021contrastive} to capture temporal structures from accelerometer and gyroscope signals for activity recognition. However, since these signals are of a similar nature, the method struggles to generalize to heterogeneous modalities such as depth-based skeleton data and IMU signals. Most critically, these approaches require individual customization during local deployment, imposing high computational demands on edge devices.

To address these intertwined challenges, we propose CLMM, a general Contrastive Learning framework for Multimodal HAR designed to operate effectively under limited labeled data regimes. Rather than directly optimizing a fully supervised model, CLMM decouples the learning paradigm into a self-supervised cross-modal alignment phase and a parameter-efficient local adaptation phase. This decoupling allows the model to leverage massive unlabeled data while remaining highly adaptable for specific dynamic environments.

Specifically, in the first stage, CLMM extracts cross-modal shared representations to overcome the semantic gap between highly heterogeneous sensors. To achieve this, we introduce an enhanced contrastive objective equipped with a hard-positive weighting mechanism, which explicitly guides the model to capture modality-invariant dynamic patterns. In the second stage, to facilitate stable customization on edge devices with minimal annotated data, we propose a primary–auxiliary collaborative training scheme guided by Exponential Moving Average (EMA)~\cite{he2020momentum}. By employing a distillation-based strategy, this scheme prevents catastrophic forgetting of the pre-trained knowledge while adaptively capturing modality-specific temporal nuances. Consequently, CLMM ensures fast convergence and robust generalization during local deployment without relying on cloud-based centralized training.
The main contributions of CLMM are summarized as follows.
\begin{itemize}

    \item We propose CLMM, a multimodal contrastive learning framework designed to bridge the gap between pre-training alignment and downstream deployment by mitigating boundary intrusion and optimization instability.
    
    \item We introduce a Hard Positive Samples Weighting mechanism to sharpen decision boundaries by dynamically emphasizing difficult instances near them.

    \item We design an EMA-guided primary--auxiliary collaborative training strategy for stable fine-tuning while mitigating catastrophic forgetting.

    \item Evaluations on three public datasets demonstrate that CLMM consistently outperforms state-of-the-art baselines and converges faster than other unsupervised learning methods.
\end{itemize}

%% file: 2.Related_work.tex
\section{RELATED WORK }
In this section, we will discuss the related studies and highlight the research gaps addressed by our proposed approach. 

\textbf{Multimodal HAR. } 
Multimodal HAR leverages collaborative sensing from multiple sensors and deep learning techniques to advance applications such as virtual reality~\cite{pascoal2019activity,rabbi2018virtual}, smart homes~\cite{adaimi2021ok,bi2017familylog}, and health monitoring~\cite{fan2021headfi,hao2013isleep}. However, training a robust HAR model usually requires a large amount of labeled data, while most sensor data is abstract and non-intuitive, making it difficult for non-experts to interpret, which in turn poses challenges for multimodal data annotation~\cite{hu2020fine,munzner2017cnn,shuai2021millieye}. Consequently, HAR research under limited labeled data has attracted widespread attention.

\textbf{Multimodal HAR with Limited Labeled Data.} To mitigate annotation scarcity, one line of work employs generative approaches. For instance, SenseGAN~\cite{yao2018sensegan} uses a semi-supervised GAN framework to generate pseudo-labeled samples. A classifier predicts labels for unlabeled data, while a discriminator distinguishes real from generated samples, effectively combining limited labels with unlabeled data. HMGAN~\cite{chen2023hmgan} learns both shared and modality-specific features via a generator, and employs a hierarchical discriminator that computes modality-level and global adversarial losses to produce multimodal samples from limited annotations. However,  GANs require training both the generator and the discriminator, making the process complex, unstable, and prone to convergence issues~\cite{jabbar2021survey,ige2022survey}.

Another approach uses self-supervised learning to learn representations directly from unlabeled data. COCOA~\cite{deldari2022cocoa} employs contrastive learning to maximize cross-modal mutual information and minimize irrelevant sample similarities. Cosmo~\cite{ouyang2022cosmo} introduces multimodal contrastive fusion to capture inter-modal consistency and complementarity. MESEN~\cite{xu2023mesen} enhances single-modality sensing via multimodal contrastive learning, while BABEL~\cite{dai2025babel} unifies multiple modalities by reformulating N-modal alignment as a set of bimodal alignment tasks. Nevertheless, these methods still struggle with hard samples, underutilized cross-modal features, and ineffective knowledge transfer in low-label settings. To overcome these limitations, we aim to develop a general HAR framework that maximizes cross-modal shared feature extraction, improves hard sample discrimination, and enables effective pretrained knowledge fine-tuning—all using limited labels—to enhance accuracy and convergence stability.

%% file: 5.System_design.tex
\begin{figure}[!t]
	\centering
	\includegraphics[width=1\linewidth]{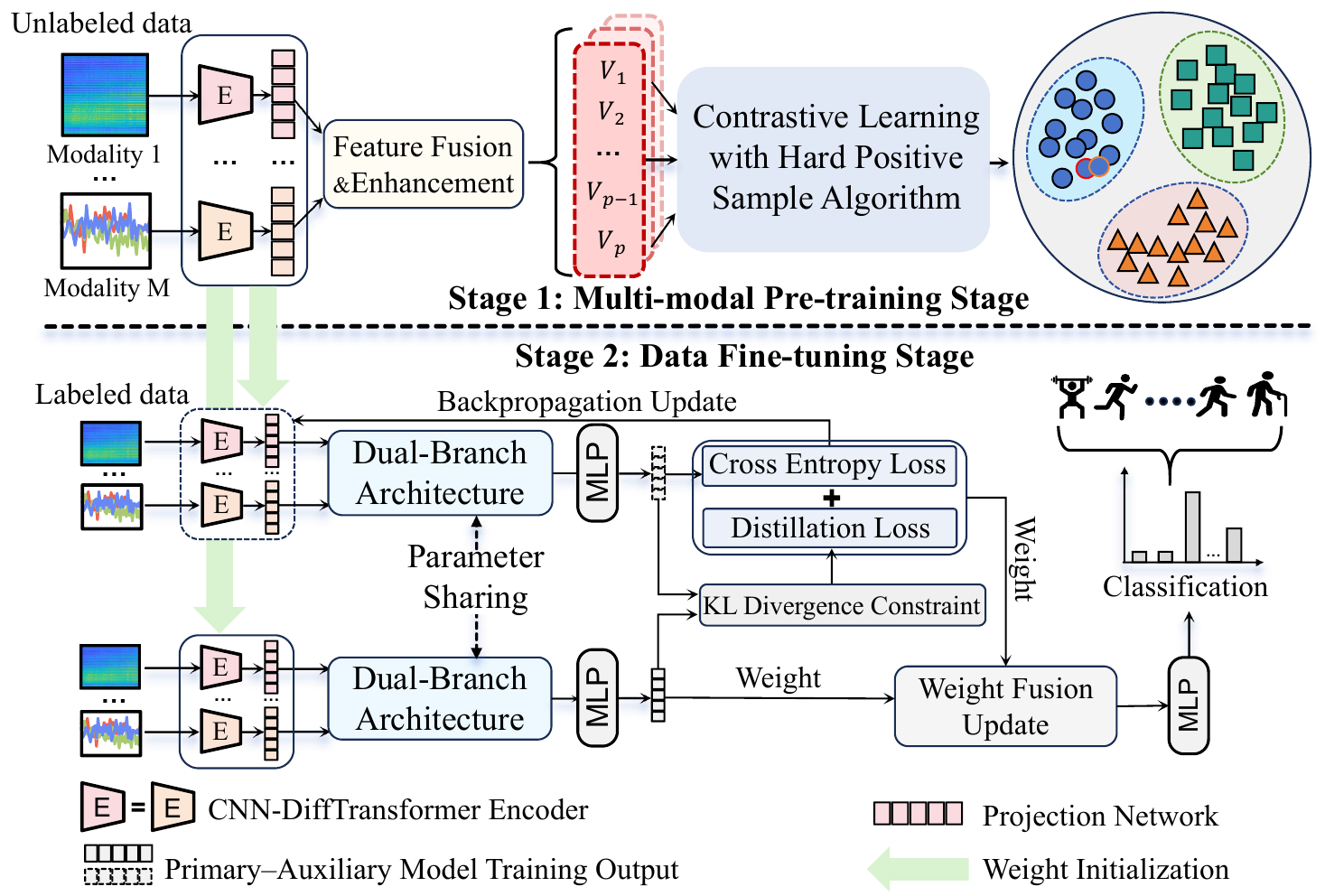}
	\caption{Overview of CLMM.}
	\label{fig5_system}
\end{figure}
\section{SYSTEM DESIGN}

This section details the design of CLMM, including multimodal contrastive fusion learning, Specific feature extraction, multimodal feature fusion, and data augmentation. The framework is illustrated in Fig.~\ref{fig5_system}.

\subsection {Multi-modal Pre-training Stage}
In the first stage, we propose a contrastive fusion learning framework to extract cross-modal shared representations from large-scale unlabeled data.
This stage consists of a CNN–DiffTransformer encoder for unimodal feature extraction, a positive–negative pair construction strategy, and a contrastive loss with hard positive weighting.

\textbf{CNN–DiffTransformer Encoder.}
Multimodal HAR data are typically heterogeneous in both structure and dimensionality.
To address this issue, we adopt a CNN–DiffTransformer encoder to extract robust unimodal features.
Given an unlabeled multimodal sample with $M$ modalities, the $j$-th modality of the $i$-th sample is denoted as $X^i_j$.
Each modality is first processed by a modality-specific CNN to extract local temporal features, yielding $F^i_j \in \mathbb{R}^{B \times S \times D}$ with $D=256$.

Since convolutional networks are limited in capturing long-range dependencies, we further employ DiffTransformer~\cite{ye2024differential} to model global temporal relationships.
Specifically, $F^i_j$ is split into $h$ parallel heads and fed into a multi-head differential attention mechanism, which suppresses noise while enhancing salient features:
\begin{equation}
\resizebox{0.9\linewidth}{!}{$
\left\{
\begin{aligned}
&[Q_1; Q_2] = F^i_j W^Q, \quad [K_1; K_2] = F^i_j W^K, \quad V = F^i_j W^V,\\
&\text{DiffAttn}(F^i_j)=\Big(
   \text{softmax}\!\left(\tfrac{Q_1 K_1^T}{\sqrt{d}}\right)
   - \lambda \,\text{softmax}\!\left(\tfrac{Q_2 K_2^T}{\sqrt{d}}\right)
   \Big)V, \\
&\text{MultiHead}(F^i_j)=\text{Concat}(\overline{\text{head}_1}, \dots, \overline{\text{head}_h}) W^o,
\end{aligned}
\right.
$}
\end{equation}
where $W^Q$, $W^K$, $W^V$, and $W^o$ are learnable parameters, and $\lambda$ is a learnable scalar initialized as $\lambda_{init}=0.8$.
The attention output is normalized to obtain the unimodal representation $Z^i_j$.

To align features across modalities, we further project each $Z^i_j$ into a shared embedding space using a modality-specific projection network:
\begin{equation}
R^i_j = \text{Norm}(\mathcal{P}_j(Z^i_j)), \quad j=1,\ldots,M,
\end{equation}
where $\mathcal{P}_j$ denotes an MLP-based projection head.
Following standard contrastive learning practice, $\mathcal{P}_j$ is used only during the contrastive pretraining stage and discarded in subsequent supervised training.

\begin{figure}[!t]
	\centering
	\includegraphics[width=5in, height=1.51in, keepaspectratio]{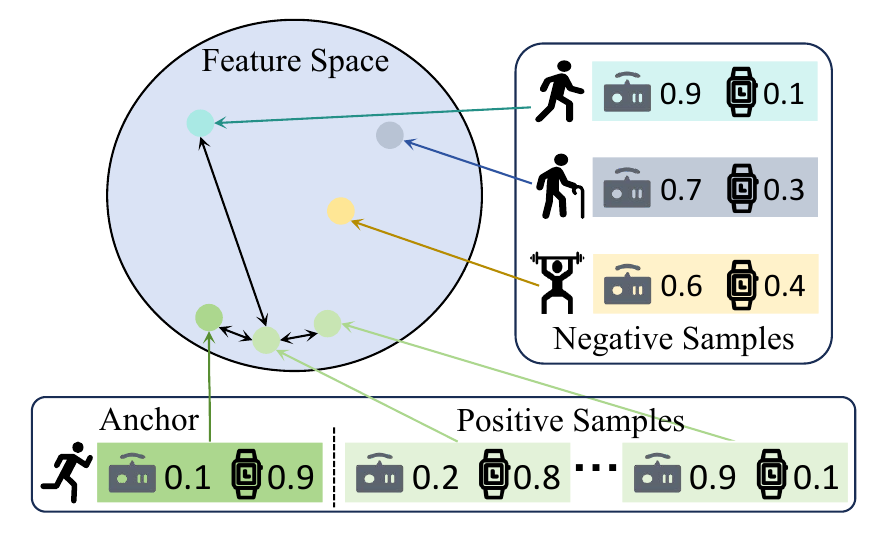}
	\caption{Fusion-based contrastive learning framework.
Positive pairs are formed by fused features from the same instance, while negative pairs are drawn from other instances in the batch. }

\vspace{4pt}
	\label{fig_sjrh}
\end{figure}

\textbf{Feature Fusion Module.}
Batch-wise contrastive learning often suffers from limited sample diversity, which motivates us to introduce a weighted feature fusion strategy.
As illustrated in Fig.~\ref{fig_sjrh}, we construct multi-scale positive and negative pairs by randomly generating multiple modality-weighted combinations, where different fusion scales are achieved by varying modality weight proportions.
Specifically, given a batch of samples $\{X^i_j \mid j=1,\ldots,M\}$, we generate $P$ modality combinations whose weights satisfy $\sum_{j=1}^{M} a_{jk}=1$.
Each combination yields a fused representation that preserves partial modality information, expressed as
\begin{equation}
V^i_k = \sum_{j=1}^{M} a_{jk} R^i_j,
\end{equation}
where $a_{jk}$ denotes the weight of the $j$-th modality in the $k$-th combination.
All fused features derived from the same sample are treated as positive samples, while those constructed from other samples in the batch are regarded as negatives.


\begin{figure}[!t]
	\centering
	\includegraphics[width=5in, height=1.5in, keepaspectratio]{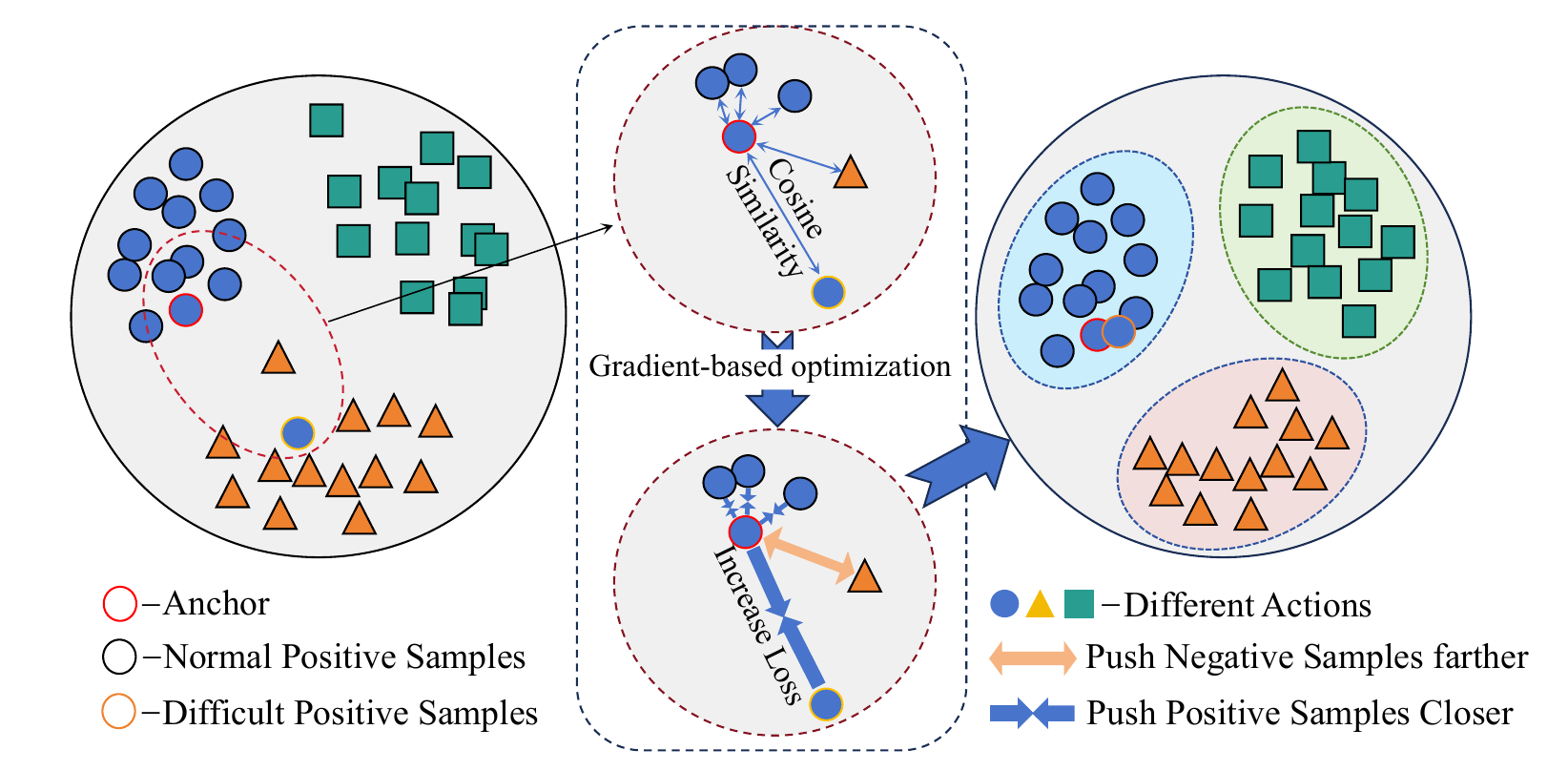}
	\caption{Workflow of the hard positive samples weighting algorithm. 
Low-similarity positive pairs (yellow) are identified as hard positives and receive stronger optimization emphasis during loss computation.}
\vspace{1pt}
	\label{fig_knzyb}
\end{figure}

\textbf{Contrastive Learning with Hard Positive Samples Weighting.}
After constructing the positive–negative sample pairs, the contrastive learning module aims to pull positive pairs closer while pushing negative pairs apart.
However, due to cross-modal heterogeneity and intra-class variations, samples from the same activity may still be far apart in the feature space.
These samples are regarded as hard positives, as they are more difficult to align with the anchor and may lead to ambiguous class boundaries.
Fig.~\ref{fig_knzyb} illustrates the workflow of the Hard Positive Samples Weighting algorithm, in which some positive samples (marked as yellow circles) deviate from their original cluster space, potentially leading to misclassification.
CLMM increases the optimization pressure on hard positive pairs during backpropagation by modulating their similarity scores in the contrastive loss, thereby guiding the model to better capture cross-modal shared information.

First, the fusion module expands a batch of $N$ raw training samples into $P \times N$ fused features.
Let $s \in S \equiv \{1, 2, \ldots, P \times N\}$ index any fused feature, and let $p$ index the other augmented features originating from the same raw sample as $s$. The feature at $s$ is the anchor and those at $p$ are positives.
Next, we compute the cosine-similarity matrix over all fused features to obtain the set of positive-pair similarities $P_{\text{positive}}$.
For each anchor, positive pairs are ranked by cosine similarity, and the lowest $\rho$ fraction is defined as the hard positive set $P_{\text{difficult}}$.
Before loss aggregation, similarities in $P_{\text{difficult}}$ are scaled by a factor $w_h < 1$, which increases their contribution to the contrastive loss and results in stronger gradients during backpropagation.
The contrastive loss is then defined as
\begin{equation}
\resizebox{0.9\linewidth}{!}{$
\mathcal{L}^{hard} = \sum_{s \in S} \frac{-1}{|P(s)|} \sum_{ p \in P(s)} \log \frac{\exp(w(s, p) \cdot \mathbf{V}_s \cdot \mathbf{V}_{p} / \tau)}{\sum\limits_{a \in S \setminus \{s\}} \exp(\mathbf{V}_s \cdot \mathbf{V}_a / \tau)}.
\label{eq:hard_contrastive_loss}
$}
\end{equation}
The weight term $w(s, p)$ is defined as follows:
\begin{equation}
w(s, p) = 
\begin{cases}
w_h, & \text{if } p \in P_{\text{difficult}} \\
1, & \text{if } p \in P_r.
\end{cases}
\end{equation}

\begin{figure}[!t]
	\centering
	\includegraphics[width=1\linewidth]{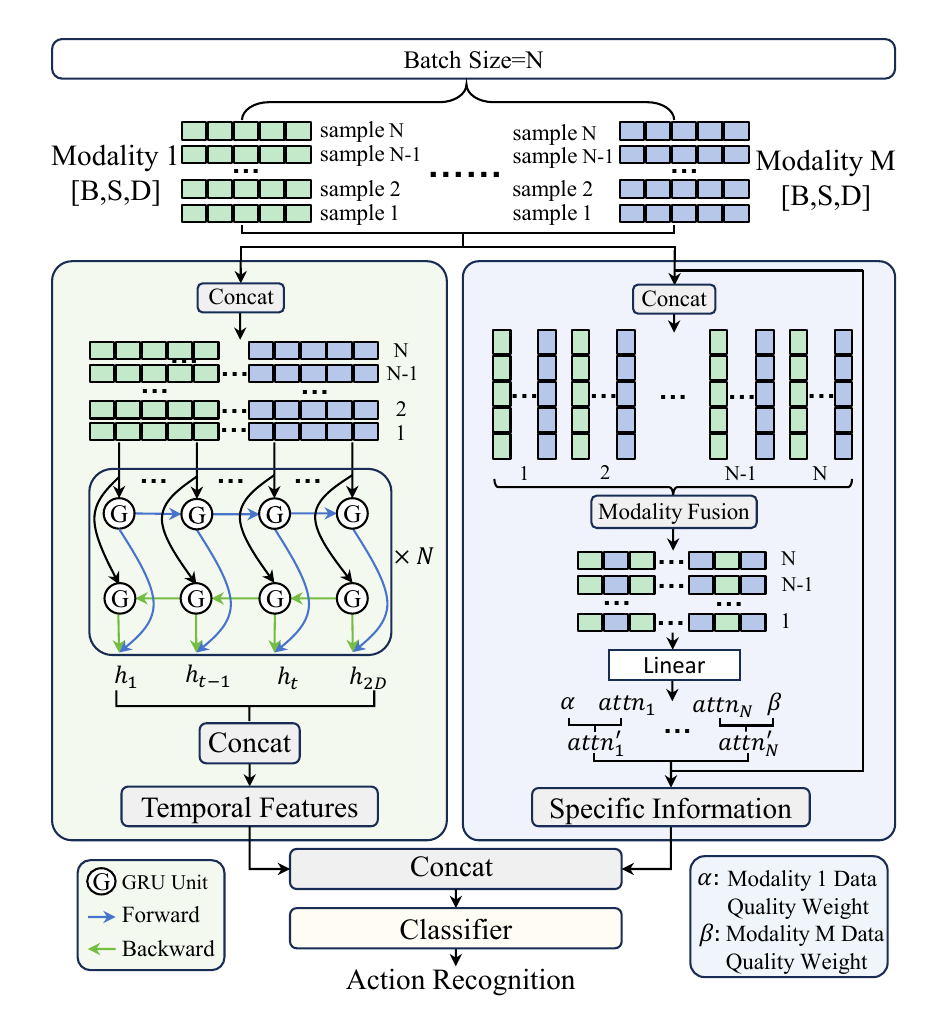}
	\caption{Dual-branch architecture extracts modality-specific information across modalities using quality-guided attention and Bi-GRU modules.}
	\label{c5_fig8}
\end{figure}

Although the similarity of hard positive pairs is scaled by a factor $w_h<1$, this operation reduces their softmax probability $q_{s,p}$. As a result, the gap between $q_{s,p}$ and the target weight $w(s,p)$ increases, leading to larger gradient magnitudes with respect to the similarity score during backpropagation.

\subsection{Data Fine-tuning Stage}
\textbf{Dual-Branch Architecture and Multimodal Feature Fusion.}
In the second stage, the model adapts pretrained shared representations to modality-specific characteristics using limited labeled data.
To this end, we adopt a dual-branch architecture (Fig.~\ref{c5_fig8}) that explicitly accounts for modality quality variations and temporal dependencies.

Let $Z_j \in \mathbb{R}^{T \times d}$ denote the feature sequence extracted by the pretrained encoder for the $j$-th modality, where $T$ and $d$ represent the temporal length and feature dimension, respectively.
To perform modality-aware fusion, a quality-guided attention mechanism assigns a fusion weight $\beta_j$ to each modality:
\begin{equation}
\beta_j = (1 - \lambda)\, \beta^{(\text{Attn})}_j + \lambda\, \beta^{(\text{QoM})}_j ,
\end{equation}
where $\beta^{(\text{Attn})}_j$ is a learnable attention weight computed from $Z_j$, and $\beta^{(\text{QoM})}_j$ is a modality quality prior estimated from unlabeled data.
The hyperparameter $\lambda$ controls the contribution of quality priors during fusion.

In parallel, a Bi-GRU is applied to the fused feature sequence to capture bidirectional temporal dependencies.
Specifically, given the input sequence $Z_j$, the Bi-GRU outputs a temporal representation $H^{\text{bi}} \in \mathbb{R}^{T \times h}$ by aggregating forward and backward hidden states.

The final fused representation is obtained by combining modality-weighted features and temporal representations:
\begin{equation}
\hat{Z} = \text{concat}(\beta_1 Z_1, \ldots, \beta_M Z_M), \quad
F = \phi(\text{concat}(H^{\text{bi}}, \hat{Z})),
\end{equation}
where $M$ denotes the number of modalities, $\phi(\cdot)$ is a multilayer perceptron, and $F$ is the output representation used for subsequent training.

\begin{figure}[!t]
\centering
\includegraphics[width=1\linewidth]{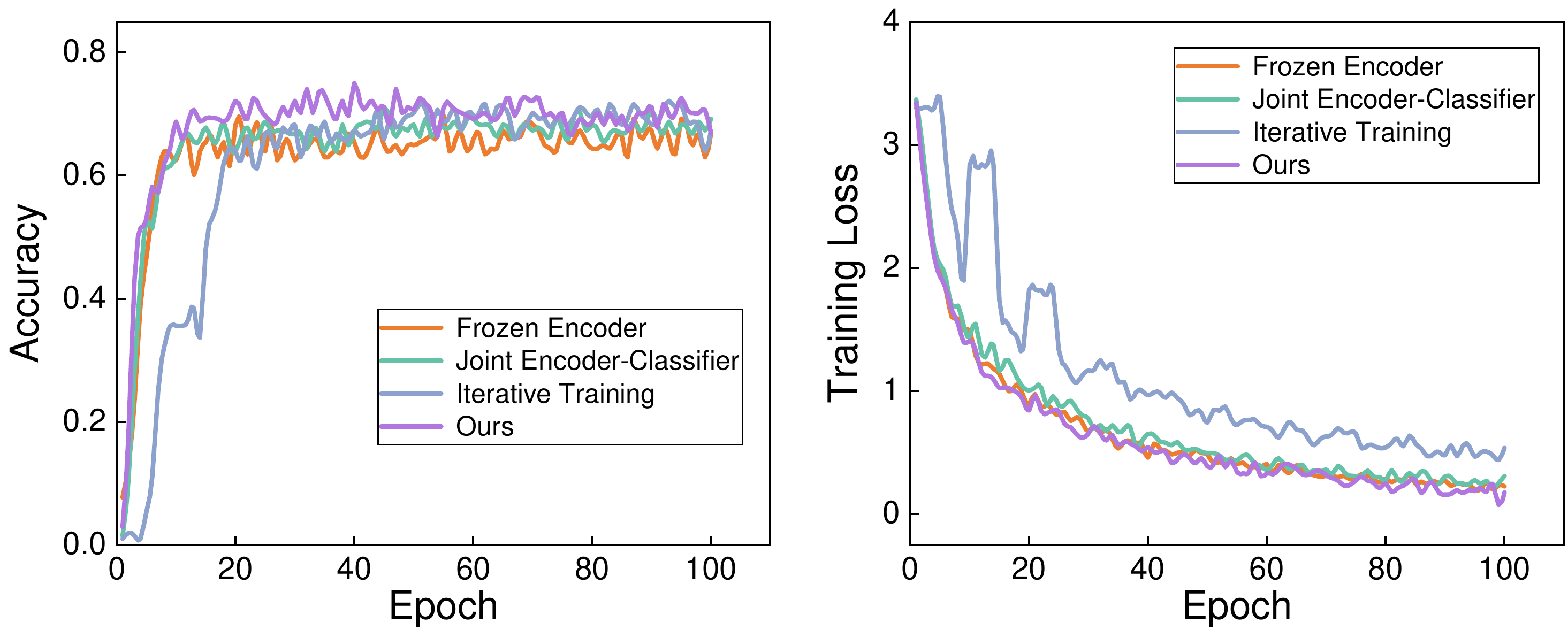}
\caption{Performance comparison between proposed Primary-Auxiliary collaborative training and other methods.}
\label{c5_fig9}
\end{figure}

\textbf{EMA-Guided Primary–Auxiliary Collaborative Training Strategy.}
Directly fine-tuning the encoder together with the classifier under limited labeled data often leads to catastrophic forgetting of pretrained shared knowledge, causing the model to degenerate into a conventional supervised fusion scheme.

To address this issue, we propose an EMA-guided Primary–Auxiliary collaborative training strategy.
As illustrated in Fig.~\ref{fig5_system}, the framework consists of a Primary model for final inference and an Auxiliary model optimized via backpropagation to learn modality-specific features.
Both models are initialized with pretrained weights from the first stage.
During training, the Auxiliary model is updated by gradient descent, while the Primary model is updated as an exponential moving average (EMA) of the Auxiliary model parameters, without receiving direct gradient updates.
This collaborative mechanism enables stable integration of shared and specific information while effectively mitigating knowledge forgetting.

The EMA update rule for the Primary model is defined as
\begin{equation}
\left\{
\begin{array}{l}
\theta_{EMA}^t = \alpha \cdot \theta_{EMA}^{t - 1} + (1 - \alpha) \cdot \theta^t, \\
\alpha = \min\left(1 - \frac{1}{t + 1}, \alpha_0\right),
\end{array}
\right.
\end{equation}
where $\theta^t$ and $\theta_{EMA}^t$ denote the parameters of the Auxiliary and Primary models at iteration $t$, respectively.
The momentum coefficient $\alpha$ is gradually increased during training to stabilize the EMA updates.

To further improve convergence, we introduce a distillation loss based on asymmetric KL divergence between the Primary and Auxiliary models.
Let $F_{\theta}$ and $F_{\xi}$ denote the outputs of the Auxiliary and Primary models, with corresponding distributions $p_{\theta} = \log \text{softmax}(F_{\theta})$ and $p_{\xi} = \text{softmax}(F_{\xi})$.
The overall training objective is defined as
\begin{equation}
\left\{
\begin{array}{l}
\mathcal{L} = \mathcal{L}_{\text{CE}} + \lambda \cdot \mathcal{L}_{\text{distill}}, \\
\mathcal{L}_{\text{distill}} = D_{\text{KL}}(p_{\theta} \parallel p_{\xi}),
\end{array}
\right.
\end{equation}
where $\lambda$ controls the contribution of the distillation loss.
As shown in Fig.~\ref{c5_fig9}, this collaborative training strategy consistently improves recognition accuracy and convergence behavior compared with the non-collaborative baseline.

\begin{table}[!t]
  \centering
  \caption{Summary of multimodal HAR datasets.}
  \renewcommand{\arraystretch}{1.3}
  \begin{tabularx}{\linewidth}{p{0.2\linewidth}cccc} 
    \hline
    \textbf{Dataset} & \textbf{Modality} & \textbf{Activity}& \textbf{Subject} & \textbf{Samples} \\
    \hline
    UTD      & IMU, Skeleton     & 27 &8   & 864    \\
    
    PAMAP2   & Acc, Gyro, Magn   & 12 &9   & 19,427 \\
    
    UTwente  & Acc, Gyro, Magn   & 13 &10  & 11,698 \\
    \hline
  \end{tabularx}
  \label{tab:dataset}
\end{table}


\subsection{Data Augmentation Module}
Data augmentation enhances the generalization capabilities of multimodal models. However, applying different strategies to modalities of the same sample may undermine cross-modal consistency, as uncoordinated operations can corrupt the semantic and temporal alignment essential for multimodal learning.

To address this, we evaluate six augmentation methods on recognition performance. Compared to the baseline without augmentation (peak accuracy 70.11\%), methods including time shifting, channel scaling, signal smoothing, and noise injection reduce accuracy (all below 68\%), mainly because they damage discriminative features. In contrast, time warping and random cropping improve performance, achieving 75.00\% and 72.12\%, respectively. The former enhances robustness by simulating variations in user movement rhythms, while the latter improves adaptation to incomplete sequences. Considering practical requirements and its superior performance, \textbf{time warping} is adopted as the primary augmentation strategy in our two-stage training.


%% file: 6.implement.tex
\section{System Implementation}

\subsection{Baselines}
This study selected six methods closely related to the CLMM as baseline models to compare their performance metrics.
\textbf{CMC~\cite{tian2020contrastive}}, \textbf{Cosmo~\cite{ouyang2022cosmo}}, and \textbf{COCOA~\cite{deldari2022cocoa}}  employ contrastive learning to align heterogeneous representations, whereas \textbf{CroSSL~\cite{deldari2024crossl}}, \textbf{STMAE~\cite{10.1145/3631415}}, and \textbf{MASTER~\cite{zhu2025master}} leverage masked modeling paradigms to address missing modalities and scalability.

\subsection{Training Configuration}


CLMM employs a two-stage training strategy for HAR.
In the pretraining stage, unlabeled multi-modal data is trained in the cloud using the SGD optimizer (learning rate 1e-2, batch size 4), with a temperature parameter $\tau=0.07$ and hard sample ratio $\alpha=0.02$.
In the fine-tuning stage, a small set of labeled data is trained locally with low computational cost, using SGD (learning rate 1e-3, momentum 0.9, batch size 4), EMA coefficient $\alpha_0=0.9$, and KL loss weight $\lambda=0.1$.

The model is implemented in PyTorch and evaluated on a Windows 11 system with an NVIDIA GeForce RTX 4060 GPU (8 GB VRAM), running PyTorch + CUDA 11.6 + cuDNN 8.8.1.
For fairness, all models are trained on identically preprocessed datasets, with each experiment repeated five times under different random initializations.
We employ three widely used metrics for evaluation: Accuracy, F1-score, and Kappa. Collectively, these metrics provide a comprehensive assessment of the overall correctness of the results, as well as the balance between precision and recall.

%% file: 7.evaluation.tex
\section{EVALUATION}

\begin{table}[!t]
  \caption{Performance on public datasets with limited labeled data. Note that “A” stands for accuracy, “F1” stands for F1-score, and “K” stands for Kappa.}
  \label{tab:experiment1}
  \renewcommand{\arraystretch}{1.3}
  \setlength{\tabcolsep}{3pt}
  \begin{tabular}{@{}>{\centering\arraybackslash}p{1.2cm} 
    *{9}{>{\centering\arraybackslash}p{0.64cm}}@{}} 
    \hline 
    \multicolumn{1}{@{}c@{}}{} & \multicolumn{3}{c}{\textbf{UTD-MHAD (5\%)}} & \multicolumn{3}{c}{\textbf{PAMAP2 (1\%)}} & \multicolumn{3}{c}{\textbf{UTwente (1\%)}} \\
    \cline{2-10} 
    \textbf{Model} & $\uparrow$ A & $\uparrow$ F1 & $\uparrow$ K & $\uparrow$ A & $\uparrow$ F1 & $\uparrow$ K & $\uparrow$ A & $\uparrow$ F1 & $\uparrow$ K \\
    \hline
    CMC    & 51.9 & 50.3 & 46.6 & 66.7 & 58.9 & 61.9 & 74.6 & 70.9 & 64.7 \\
    COSMO  & 62.2 & 57.7 & 58.4 & 69.6 & 61.2 & 66.4 & 72.6 & 68.5 & 63.2 \\
    COCOA  & 58.1 & 54.3 & 52.4 & 71.5 & 67.7 & 64.3 & 67.2 & 64.2 & 62.7 \\
    CroSSL & 61.5 & 59.4 & 58.7 & 72.6 & 71.1 & 72.5 & 73.5 & 69.6 & 67.4 \\
    STMAE & 62.9 & 51.5 & 58.1 & 61.6 & 52.2 & 55.9 & 77.9 & 66.3 & 73.2 \\
    MASTER & 69.3 & 66.7 & 65.2 & 76.2 & 73.2 & 73.3 & 81.5 & 74.3 & 77.4 \\
    \hline 
    CLMM   & \textbf{75.0}  & \textbf{73.3} & \textbf{73.9} & \textbf{85.2} & \textbf{82.4} & \textbf{83.4} & \textbf{88.5} & \textbf{86.1} & \textbf{85.3} \\
    \hline
  \end{tabular}
\end{table}

In this section, we evaluate the performance of CLMM under various conditions and compare it against baseline models to assess its overall effectiveness. 

\subsection{Datasets}
We evaluate CLMM on three public datasets covering diverse modalities and scenarios, as summarized in Table~\ref{tab:dataset}.

\textbf{UTD~\cite{chen2015utd}.}
A comprehensive dataset containing synchronized RGB, depth, skeleton, and IMU data. To prevent overfitting, we strictly utilize the lower-dimensional skeleton (30Hz) and IMU (50Hz) modalities. Data from 6 subjects are used for training and 2 for testing.

\textbf{PAMAP2~\cite{reiss2012introducing}.}
PAMAP2 records 9 subjects performing 12 activities. Accelerometer, gyroscope, and magnetometer signals (100Hz) were collected from chest, wrist, and ankle sensors. We use these three modalities, with 7 subjects for training and 2 for testing.

\textbf{UTwente~\cite{shoaib2016complex}.}
This dataset involves 10 subjects and 13 activities, collected by smartwatch and smartphone sensors. It includes accelerometer, gyroscope, linear acceleration, and magnetometer (50Hz). 
Data from both devices are fused to construct multi-dimensional feature vectors, and the dataset is split 8:2 for training and testing.

\subsection{Performance on Different Datasets}

\begin{figure}[!t]
\centering
\includegraphics[width=1\linewidth]{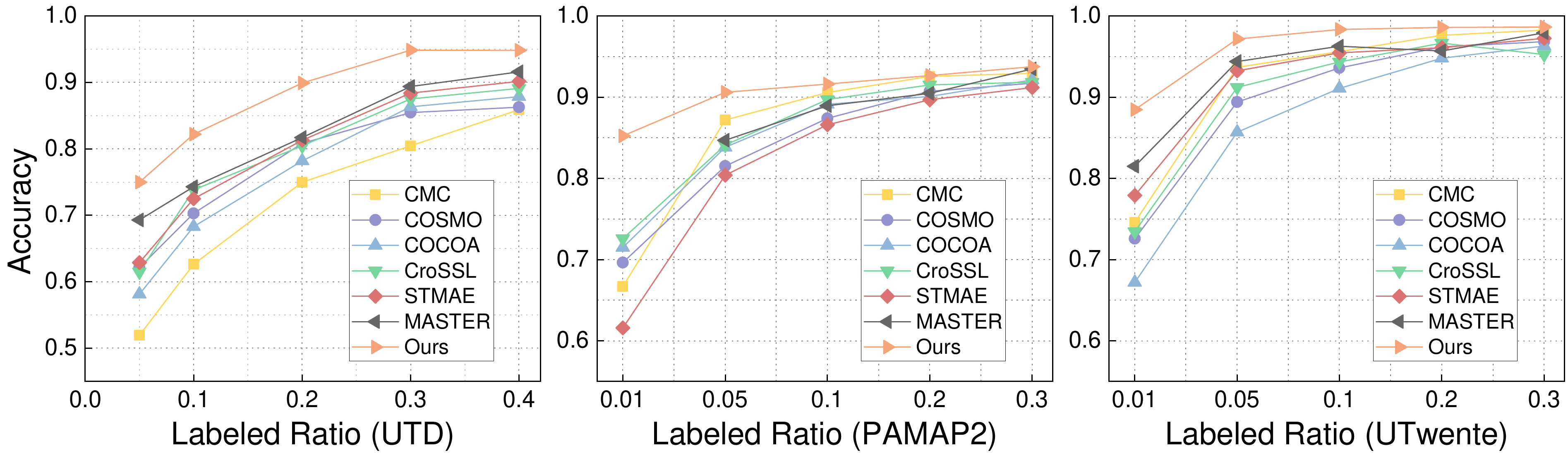}
\caption{Accuracy comparison with different amounts of labeled data. 
         CLMM consistently outperforms baselines 
         and achieves CMC’s  full-data equivalence with $\leq 20\%$ labels.}
\label{fig_labeled}
\end{figure}
 \begin{figure}[!t]
\centering
\includegraphics[width=1\linewidth]{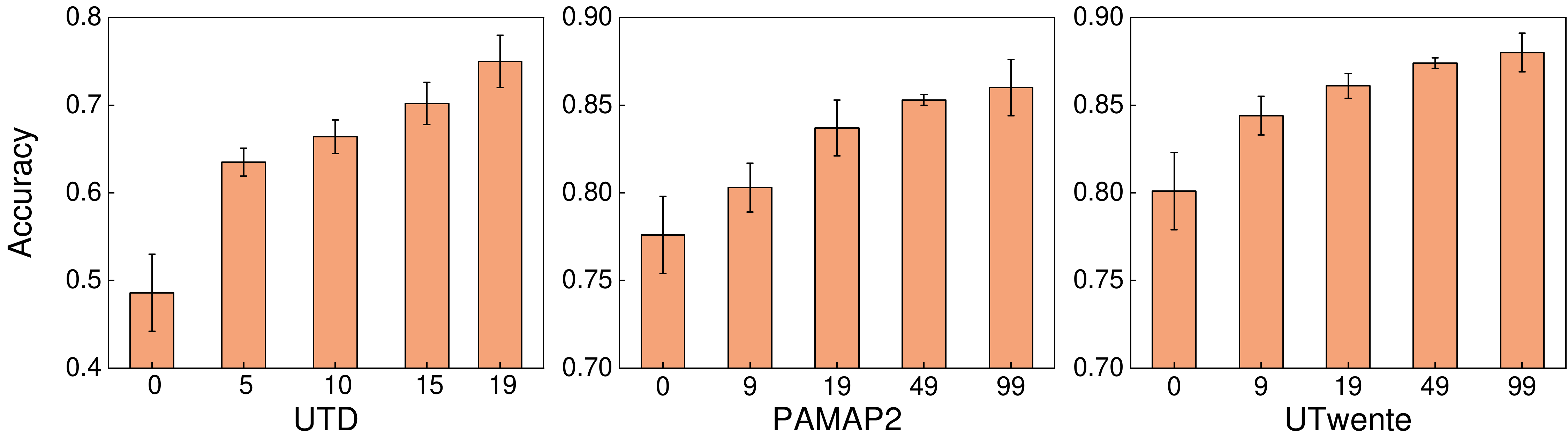}
\caption{Accuracy of CLMM with different amounts of unlabeled data. Values in X-axis are $\frac{\#\text{Unlabeled data}}{\#\text{Labeled data}}$.}
\label{fig_unlabeled}
\end{figure}


\textbf{Performance Under Limited Labeled Data.}
We evaluate different methods under limited labeled data to simulate realistic data-scarce HAR scenarios.
Labeling ratios of 5\%, 1\%, and 1\% are adopted on three datasets, corresponding to 27, 93, and 155 labeled samples, respectively.
As shown in Table~\ref{tab:experiment1}, CLMM consistently achieves the best performance across all datasets, outperforming COSMO by 12.85\%, 15.0\%, and 15.9\%.
Early contrastive methods such as CMC and COSMO exhibit limited robustness under low-label settings, while more recent approaches improve performance but still degrade when supervision becomes extremely sparse.
These results demonstrate that CLMM effectively benefits from hard positive mining and EMA-guided collaborative training, enabling stable knowledge transfer and robust discrimination under limited supervision.




\textbf{Impact of Different Amounts of Labeled Data.}
Fig.~\ref{fig_labeled} compares recognition performance under varying amounts of labeled data.
Although all methods improve with increased supervision, CLMM consistently outperforms the baselines, with a more pronounced advantage in extremely low-label regimes.
For example, on UTD with only 27 labeled samples, CLMM exceeds CMC, COSMO, COCOA, CroSSL, STMAE, and MASTER by 23.07\%, 12.85\%, 16.88\%, 13.5\%, 12.1\%, and 5.7\% accuracy, respectively.
Moreover, CLMM achieves performance comparable to fully supervised CMC using only 15\%, 10\%, and 20\% labeled data on UTD, UTwente, and PAMAP2, highlighting its label efficiency.
This advantage arises from the primary–auxiliary collaborative training that effectively integrates shared and modality-specific information.

\begin{figure}[!t]
\centering
\begin{minipage}{0.32\linewidth} 
    \centering
    \setlength{\abovecaptionskip}{6pt}
\setlength{\belowcaptionskip}{-10pt}\includegraphics[width=\linewidth]{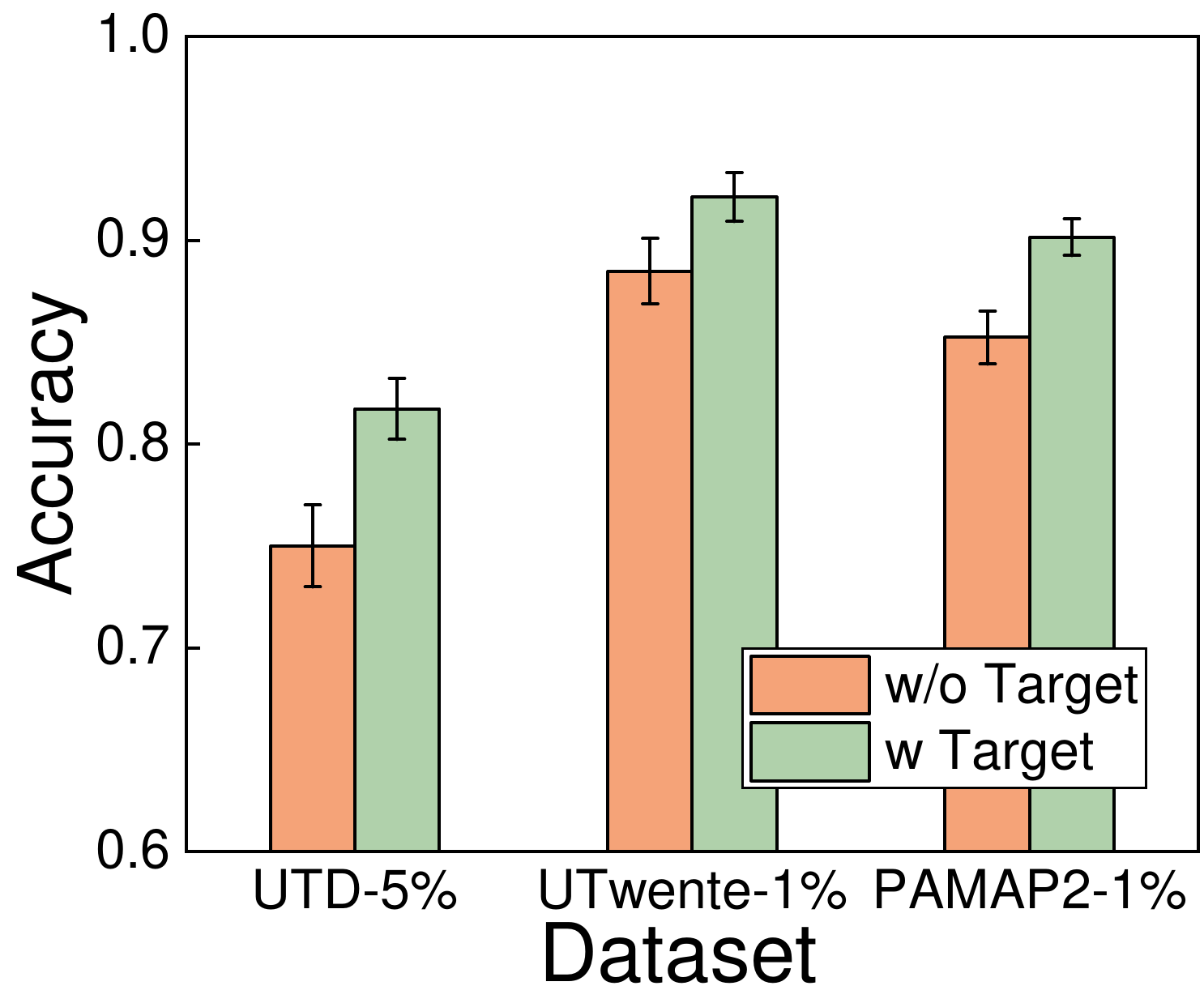}
    \caption{Accuracy wi\-th/without labeled d\-ata from target subjects.}
     \label{fig_datasource}
\end{minipage}
\begin{minipage}{0.32\linewidth} 
    \centering
    \setlength{\abovecaptionskip}{6pt}
\setlength{\belowcaptionskip}{-10pt}\includegraphics[width=\linewidth]{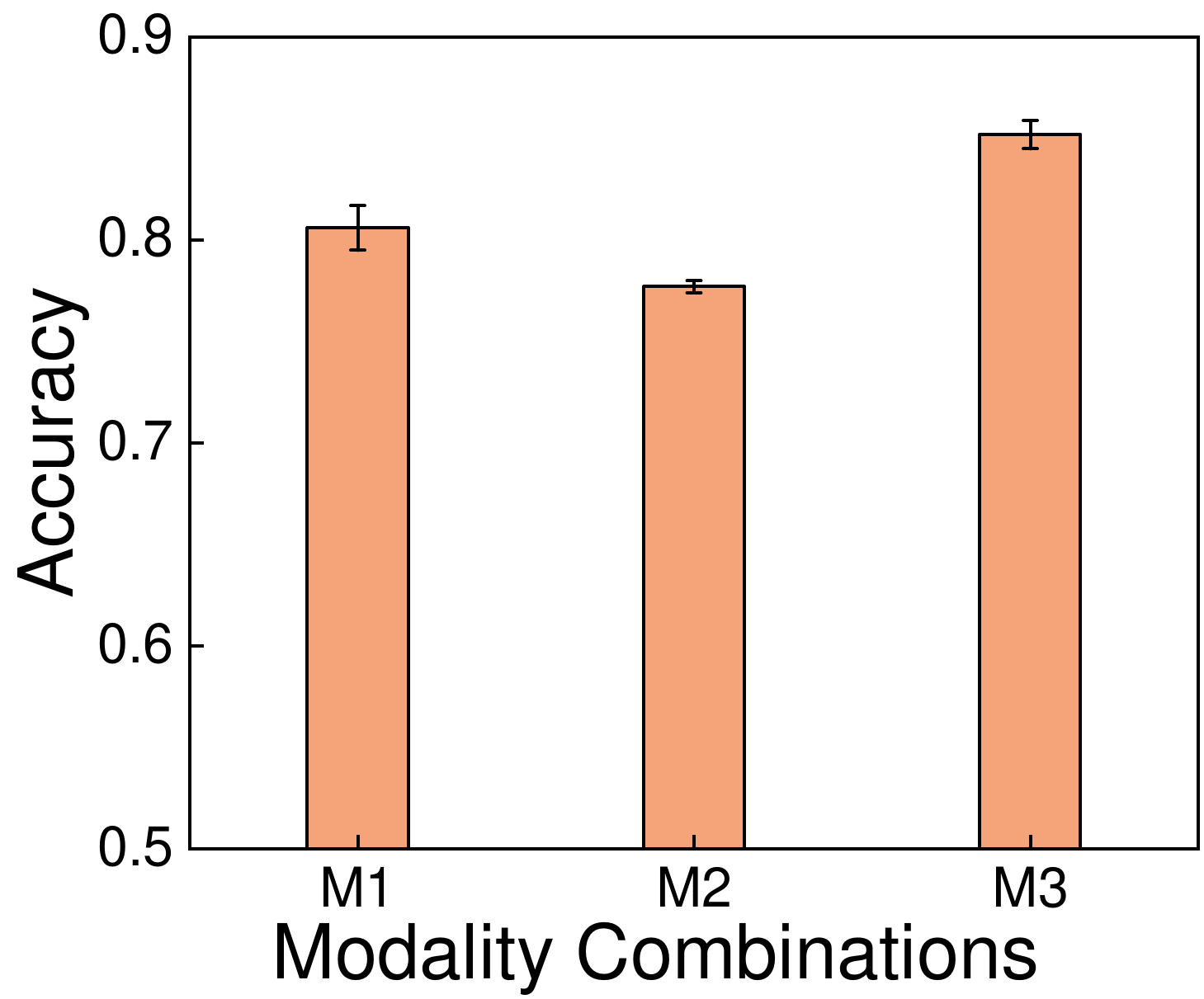}
    \caption{Three sensor c\-ombinations on PAMA\-P2.}
    \label{fig_mtzh}
\end{minipage}
\begin{minipage}{0.32\linewidth} 
    \centering
    \setlength{\abovecaptionskip}{6pt}
\setlength{\belowcaptionskip}{-10pt}\includegraphics[width=\linewidth]{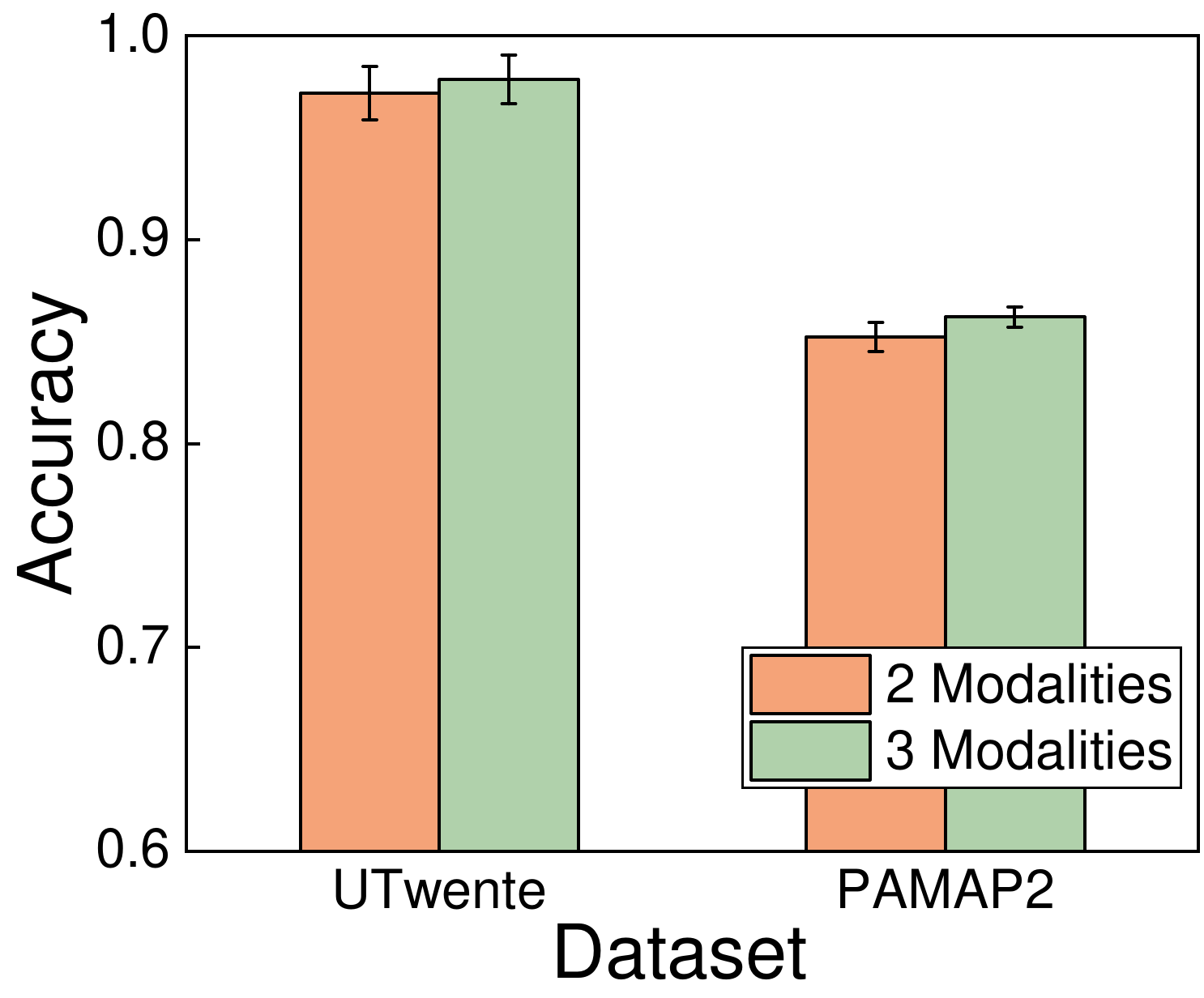}
    \caption{Dual-to-triple m\-odality expansion on U\-Twente and PAMAP2.}
   \label{fig_mtkz}
\end{minipage}
\end{figure}

\begin{figure}[!t]
    \centering
\includegraphics[width=1\linewidth]{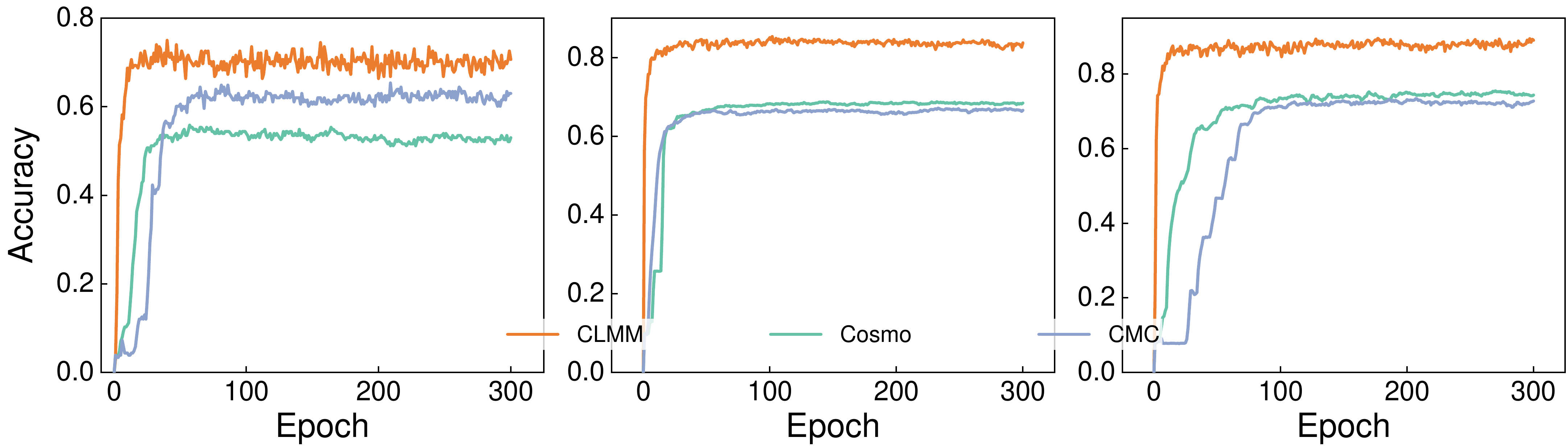}
	\caption{Convergence Performance of CLMM, CMC and Cosmo on the UTD, PAMAP2, and UTwente Datasets.}
	\label{fig_convergence}
\end{figure}

\textbf{Impact of Different Amounts of Unlabeled Data.}
We further fix the labeled set size and evaluate the effect of varying amounts of unlabeled data.
As shown in Fig.~\ref{fig_unlabeled}, recognition accuracy consistently improves as more unlabeled data are introduced, confirming CLMM’s ability to exploit cross-modal shared information.
When the unlabeled data scale becomes sufficiently large, performance gains gradually saturate, likely due to noise accumulation.

\begin{figure*}[!t]
\centering
\includegraphics[width=1\linewidth]{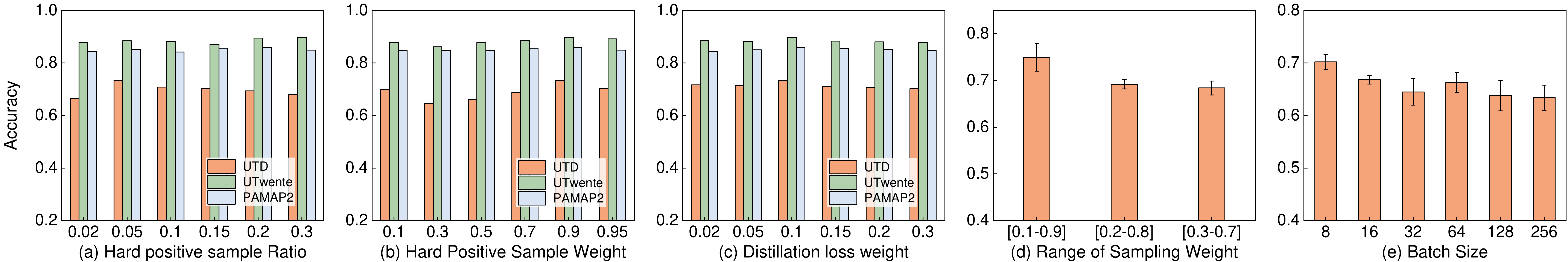}
\caption{Hyperparameter sensitivity analysis of CLMM across three datasets.}
\label{fig_超参数}
\end{figure*}

\textbf{Impact of Data Source.}
In the previous experiments, the training and testing data came from different users, resulting in a large domain gap. Fig.~\ref{fig_datasource} compares CLMM’s performance when the second-stage training data are drawn from the test users (w. target) versus from other users (w/o target). The results show that CLMM achieves significantly higher accuracy when the labeled training data comes from the test users. For example, on the UTD multimodal dataset, CLMM reaches an average accuracy of 85.23\% with only 27 labeled samples from the test users.

\subsection{Impact of Modalities on Performance}

\textbf{Impact of Different Modality Combinations.}\label{exp:combination}
We compare three sensor combinations: (1) accelerometer + magnetometer, (2) gyroscope + magnetometer, and (3) accelerometer + gyroscope.
As shown in Fig.~\ref{fig_mtzh}, the accelerometer–gyroscope combination achieves the best performance, as it jointly captures motion intensity and postural dynamics, yielding more discriminative representations.
In contrast, magnetometer-based combinations perform worse due to their limited sensitivity to common activities and susceptibility to environmental interference, which introduces additional noise.

\textbf{Impact of Modality Expansion.}
We further evaluate the effect of modality expansion on UTwente and PAMAP2.
As shown in Fig.~\ref{fig_mtkz}, expanding from dual to triple modalities improves accuracy to 97.8\% and 86.2\%, respectively, confirming that incorporating complementary sensor information enhances recognition performance.



\begin{table}[t]
\centering
\caption{Efficiency comparison on PAMAP2 Dataset.}
\label{tab:efficiency}
\begin{tabular}{lccccc}
\toprule
Model &Acc (\%) & Params (M) & FLOPs (G) & Latency (ms)   \\
\midrule
CMC &66.7 & 19.83 & 0.59 & 2.238  \\
COSMO &69.6 & 19.83 & 1.04 & 2.523  \\
COCOA &71.5 & 15.47 & 0.65 & 2.834  \\
CroSSL &72.6 & 17.79 & 1.68 & 4.172  \\
STMAE &61.6 & 23.44 & 1.14 & 2.228 \\
MASTER &76.2 & 21.37 & 1.79 & 8.353  \\
CLMM &85.2 & 25.48  & 1.46 & 3.838  \\
\bottomrule
\end{tabular}
\end{table}

\subsection{System Performance}

\textbf{Convergence Performance.}
The second-stage on-device training of CLMM requires fast convergence under limited labeled data.
We evaluate convergence on three multimodal datasets, as shown in Fig.~\ref{fig_convergence}.
Compared with CMC and COSMO, CLMM converges significantly faster, typically within 20–40 epochs.
This advantage stems from large-scale unlabeled pretraining in the first stage and the collaborative training strategy that effectively exploits limited labeled data in the second stage.

\textbf{Inference Performance.}
We further evaluate inference efficiency on edge devices using the PAMAP2 dataset.
As reported in Table~\ref{tab:efficiency}, lightweight methods such as CMC and COSMO achieve low latency but yield accuracies below 70\%.
Although MASTER attains higher accuracy, its high computational cost (1.79 GFLOPs) and inference latency (8.353 ms) hinder edge deployment.
In contrast, CLMM achieves 85.2\% accuracy with a latency of 3.838 ms ($54\% \downarrow$) and lower computational cost than MASTER.
Despite a slightly larger parameter size, CLMM does not incur noticeable latency overhead, making it well suited for resource-constrained edge scenarios.
\begin{table}[!t]
\caption{Ablation Experiment Analysis of CLMM Components.}
\label{tab:ablation}
\centering
\begin{tabularx}{\linewidth}{ 
  >{\centering\arraybackslash}X 
  >{\centering\arraybackslash}X 
  >{\centering\arraybackslash}X 
  >{\centering\arraybackslash}X 
  >{\centering\arraybackslash}X 
  >{\centering\arraybackslash}p{2cm}   
}
\toprule
ST & M1 & M2 & M3 & M4 & Accuracy (\%) \\
\midrule
$\times$ & $\times$ & $\times$ & $\times$ & $\times$ & 60.9 \\
$\checkmark$ & $\times$ & $\times$ & $\times$ & $\times$ & 63.4  \\
$\times$ & $\checkmark$ & $\times$ & $\times$ & $\times$ & 67.1  \\
$\times$ & $\checkmark$ & $\checkmark$ & $\times$ & $\times$ & 69.1  \\
$\times$ & $\checkmark$ & $\checkmark$ & $\checkmark$ & $\times$ & 72.2  \\
$\times$ & $\checkmark$ & $\checkmark$ & $\checkmark$ & $\checkmark$ & 75.0    \\
\bottomrule
\addlinespace[0.5ex]
\multicolumn{6}{l}{\footnotesize $\checkmark$: The component is used; $\times$: The component is not used.}
\end{tabularx}
\end{table}

\textbf{Ablation Study.}
We conduct ablation studies to assess the contribution of four core components: the CNN-DiffTransformer encoder (M1), dual-branch architecture (M2), hard positive sample weighting (M3), and collaborative training (M4).
In addition to the standard with/without M1 comparison, the CNN-DiffTransformer is replaced by a standard Transformer to isolate the effect of differential attention.
As shown in Table~\ref{tab:ablation}, recognition performance improves consistently with progressive module integration, confirming the effectiveness of each component.

\subsection{Impact of Hyperparameters}

We evaluate CLMM’s sensitivity to key hyperparameters in hard positive mining, distillation regularization, and feature fusion, as shown in Fig.~\ref{fig_超参数}. Overall, all parameters exhibit stable and interpretable trends, allowing reliable configuration across datasets.

For hard positive mining (Fig.~\ref{fig_超参数}(a)-(b)), the optimal hard positive ratio depends on modality heterogeneity.
UTD, with pronounced cross-modal differences, performs best with a small ratio (0.02), whereas UTwente and PAMAP2, dominated by homogeneous IMU modalities, favor larger ratios (0.3 and 0.2).
In contrast, the hard positive weight shows consistent behavior across datasets, with performance steadily increasing and peaking at 0.9, highlighting the importance of emphasizing informative hard positives.

For distillation-based regularization (Fig.~\ref{fig_超参数}(c)), all datasets achieve optimal performance at a distillation weight of 0.1, indicating that moderate regularization effectively transfers knowledge without hindering the contrastive objective.

We further analyze fusion strategy and batch size (Fig.~\ref{fig_超参数}(d)-(e)).
A wider modality-weight sampling range ([0.1, 0.9]) significantly outperforms a narrower range ([0.3, 0.7]), achieving 75\% versus 68.4\% accuracy on UTD, which underscores the benefit of diverse modality combinations.
Finally, CLMM consistently favors smaller batch sizes, as the feature fusion module implicitly enriches positive and negative samples within each batch, making large batches unnecessary and even detrimental due to increased optimization difficulty.

%% file: 9.conclusion.tex
\section{CONCLUSION} 
We propose CLMM, a low-cost and generalizable framework for multimodal human activity recognition based on contrastive learning. CLMM employs a two-stage training strategy that leverages large-scale unlabeled data on the cloud and limited labeled data on the edge. By integrating a CNN-DiffTransformer enhanced with hard-positive sample weighting and a dual-branch architecture, CLMM can effectively extract both shared cross-modal information and modality-specific features. Finally, we adopt a primary–auxiliary collaborative training strategy to fuse shared and specific representations, enabling stable fine-tuning and inheritance of pre-trained knowledge. Experimental results show that CLMM achieves robust multimodal HAR even with limited labeled data.